# Learning, Social Intelligence and the Turing Test
## – *why an "out-of-the-box" Turing Machine will not pass the Turing Test*

Bruce Edmonds and Carlos Gershenson

Centre for Policy Modelling, Manchester Metropolitan University
bruce@edmonds.name
Departamento de Ciencias de la Computación, Instituto de Investigaciones en Matemáticas Aplicadas y en Sistemas, Universidad Nacional Autónoma de México
cgg@unam.mx

**Abstract**. The Turing Test (TT) checks for human intelligence, rather than any putative general intelligence. It involves repeated interaction requiring learning in the form of adaption to the human conversation partner. It is a macro-level post-hoc test in contrast to the definition of a Turing Machine (TM), which is a prior micro-level definition. This raises the question of whether learning is just another computational process, i.e. can be implemented as a TM. Here we argue that learning or adaption is fundamentally different from computation, though it does involve processes that can be seen as computations. To illustrate this difference we compare (a) designing a TM and (b) learning a TM, defining them for the purpose of the argument. We show that there is a well-defined sequence of problems which are not effectively designable but are learnable, in the form of the bounded halting problem. Some characteristics of human intelligence are reviewed including it's: interactive nature, learning abilities, imitative tendencies, linguistic ability and context-dependency. A story that explains some of these is the Social Intelligence Hypothesis. If this is broadly correct, this points to the necessity of a considerable period of acculturation (social learning in context) if an artificial intelligence is to pass the TT. Whilst it is always possible to 'compile' the results of learning into a TM, this would not be a designed TM and would not be able to continually adapt (pass future TTs). We conclude three things, namely that: a purely "designed" TM will never pass the TT; that there is no such thing as a general intelligence since it necessary involves learning; and that learning/adaption and computation should be clearly distinguished.

## 1. Introduction

The approaches in Turing's two most famous papers contrast markedly. The definition of a computation, in the form of a Turing Machine (TM), is a micro-level specification of a device (its design and rules for operation) from which computable functions can be defined [22]. It specifies what happens when a TM that has been built is set going. It is a formal definition that defines the set of computable functions. The Turing Test (TT) is a macro-level test that is applied to an existing entity that is "running" [23]. It is not formally defined but a practical test, intended to be feasible to implement. Here intelligence is not something to be proved but demonstrated.

As pointed out by French [10], the TT is not a test of a putative "general intelligence" but a test of a specific kind of intelligence – normal human intelligence. There may well be intelligent entities that might not pass the TT, for example a human suffering from influenza or an alien whose language we do not know. The point of the TT is that if some entity passes it then it is hard to deny that this entity is intelligent – it short-cuts possible quibbling, and thus opens up the possibility that an artificial entity could be judged as intelligent.

The TT consists of a conversation over a period of time between a tester and the entity being tested. This requires an ability to learn or adapt to what the tester has said, including: the topic of conversation, the style, the detected context that the tester is coming from, and the importance given to particular issues. Clearly the TT is harder the longer it goes on for. It is far easier to fool someone if one only talks to them for a limited period of time (using, for example, rote learned scripts) than if there is time for topics to be revisited with more testing questions, checking consistency with what went before as well as common knowledge and assumptions. It is the interactive and adaptive nature of the TT that makes it so challenging, revealing shallow strategies (such as simplistic syntactic approaches) as inadequate[1].

The arguments in this paper rest upon the difference between computation (defined by a halting TM) and adaption, which is an essential part of the intelligence that is tested for by the TT. Thus in section 2 we argue that computation and adaptivity are different kinds of things, giving an example where they can be shown to differ. We then briefly review some of the characteristics of human intelligence in section 3 and look at some of the consequences in terms of passing the TT using purely designed TMs in section 4. We then consider the broader nature of intelligence and its role in section 5 before concluding.

## 2. Learning vs. Computation

The relationship between adaptive processes, i.e. learning in the broadest sense, and computation is not straightforward. Clearly learning involves processes that could sensibly be characterized as a computation, as the field of Machine Learning amply demonstrates. Similarly, some computations – when acting upon some internal data structure and outputting an updated data structure in response to new information – might be sensibly thought of as a learning process. The physical device of "a computer" can clearly be set up to do both learning and computation[2]. Thus the question arises whether they do, fundamentally, differ. In particular it is sometimes assumed that learning processes are just a particular kind of computation. The core of our argument here is that the TM (or equivalent) is not an *adequate* model of learning processes, and hence misses out a crucial aspect of intelligence, its adaptivity.

---

[1] Here we assume that the TT is conducted over a suitably extended period of time, since this tests intelligence as we know it, what is called the "Long Term Turing Test" in [8].

[2] Though it is interesting that the first use of the word "computers" referred to people – for example the women that did calculations as part of the effort to break codes during WWII at Bletchley Park. The word seems to have been transferred to the devices that took over this task when they were constructed [1].

Many different ways of defining the computable functions turned out to be the same, resulting in the "Church-Turing" thesis that these *were* the class of effective computations, including observed processes in the real world [5]. It also meant that the details of the computation processes were not deemed as crucial, but rather what was, and was not, finitely computable. Turing also showed that there was a Universal TM, one that could be given a program number effectively executed that program. Thus, in a deep sense there is a universal characterization of computation. An observed, physical process is meaningfully characterized as a computation if its inputs compared to outputs can be effectively predicted (at least at the micro, step-by-step level) by a computable function (via a suitable mapping).

There is no such agreement on the definition of a learning process. Rather there are many different kinds of learning process, each with different properties and assumptions. Indeed the "No Free Lunch" theorems [25] from Machine Learning show (in an ultimate, abstract sense) that no learning algorithm is better than any other. In other words, that to find a better learning algorithm you need to use some prior knowledge about the class of situations within which the adaption is taking place (in ML terms this is characterized as the search problem), which means that the class of situations being learned about is a proper subset of all possible situations. This indicates that there is no universally best learning algorithm, however clever the learning strategy is (e.g. using meta-reasoning or learning, or lots of special cases).

Thus, for the purposes here, we will define a kind of learning process, an *adequate incremental learner*, as follows:

- As a computational process (e.g. a TM) plus a model, which is a set of data;
- Where the data can be judged as to its truth or adequacy about something exterior, which we shall call the learning "target" (for example via another TM that produces a prediction, or output, from the model with respect to the target);
- Where the computation is iteratively provided with information from the target such that during each iteration the data is modified by the TM using the information so that the updated model is at least as adequate as before[3];
- After a finite number of iterations the model becomes maximally adequate.

This is clearly not a universal definition of learning, since it excludes known learning processes (e.g. ones that sometimes degrade the model). However, it is also clearly *a* learning process and will thus do for the purposes here. Here we use it to show that there is something that can be learned (in the above sense) that cannot be computed (by a TM). The particular problem we will use for this demonstration is that of finding the lookup table (or TM) that solves the "Limited Halting Problem".

The "Limited Halting Problem" is a sequence of increasingly difficult problems, indexed by integer, n, and defined as follows. Let us assume there an enumeration of TMs, $\{P_1, P_2, P_3 \ldots\}$. The limited halting problem is that of "given *n*, does $P_i(j)$ eventually halt where $i,j \leq n$". Let us call this sequence of problems $\{H_1, H_2, H_3, …\}$, where $H_n(i,j)$ is 1 if $i,j \leq n$ and $P_i(j)$ eventually halts, 0 if $i,j \leq n$ and $P_i(j)$ does not halt, and undefined otherwise. Each problem $H_i$ *is* computationally decidable, since it can be implemented as a simple 2D look-up table with the rows *{1, … n}* for possible program indices, $P_i$, and columns *{0, … n−1}* for possible inputs, *j*, and entries 0 or 1 depending on whether $P_i(j)$ eventually halts. The problem is not the existence of this table but of finding the right entries for it. The definition of computability is not constructive, it is sufficient that there *exist* a program to compute a function, not that we can find or implement this program.

---

[3] One can see the TM as presented with an index representing the set of information: new information from the target, and the present model and outputting an index representing the updated model (which would replace the old version of the model).

The point is that there *is* an adequate incremental learner that can learn the lookup table for $H_n$ given any n but there is *no* TM that can implement this lookup table (or equivalent TM) given any n. We now give an informal proof of each part.

The following non-terminating algorithm establishes that, given an integer *n* the lookup table that solves $H_i$ can be learnt by an adequate incremental learner.

```
Build a n×n table with all entries 0
s := 1
Repeat for ever
    For i from 1 to n
      For j from 0 to n-1
        Calculate Pi(j) for s steps
        If it has terminated
          then change entry at (I,j) to 1
      Next j
    Next i
```

Any $P_i(j)$ that halts will do so after a finite period of time, so eventually the table being adapted by the above algorithm will have the correct entries for solving $H_n$ although one might well not know when one gets to this point and there is no effective method for knowing whether one has (otherwise we could solve the general halting problem). This algorithm fulfills the definition above.

Now to show that there is no effective method for, given *n*, of finding the TM that implements the solution to $H_n$. Suppose there were a computable function *f(x)* such that $P_{f(n)}$ computes $H_n$. In other words, of effectively finding the program index that implements the table for $H_n$, (which we know exists). This is equivalent to having a general and effective procedure for constructing the TM for $H_n$ given *n*.

Suppose there was such a computable function, *f(x)*, then to decide whether $P_i(j)$ halts, calculate $P_{f(max(i,j))}(i,j)$. This is defined since $i,j \leq max(i,j)$. See if the answer is *1* or *0*. $P_{f(n)}$ is computable via the universal TM [5]. Thus if *f(x)* was computable we could effectively solve the general halting problem, which we know is impossible [22]. Thus *f(x)* is not computable, that is to say there is no TM that computes it.

Thus there is a learning process whose "resulting" model is not computable by a TM. Learning is different from computation. Of course, in a way, this is obvious since a computation is, as defined, not a process but a formally defined function (albeit possibly defined using a process), whilst learning is an on-going process. Learning can be seen as a kind of non-predefined change in a computational process [13]. A TM cannot implement this since the change is not predefined.

Formalists may well be dissatisfied with the above demonstration since the TM has to halt whilst the adequate incremental learner does not. However this difference is at the crux of the matter. Computation, as defined by a halting TM, is not a process but the *result* of a process – the TM is merely a means of defining which functions are computable[4]. A possible confusion might arise if people conflate what we call a "computer" (the physical object we use) and what is formally defined as "computable" (using a halting TM or other definition). It is true that the intermediate state of any process that implements a complete computation could, itself, be seen as a computation of that intermediate state, but that intermediate state is not part of the definition of the complete computation[5]. This difference becomes important when we are considering what sort of entity could pass the TT.

## 3. Human Intelligence

We briefly consider some of the characteristics of human intelligence, since the TT tests for an ability that results from human intelligence: the ability to converse. These include being able to:
- continually react to social signals
- imitate and learn from others
- detect what is the appropriate social context
- react in ways appropriate to the detected social context
- imagine what it is like to be other people
- use other people for sensing and filtering relevant information from the environment
- use other people to act on behalf of ourselves
- make alliances and friendships, maintained by frequent interaction
- acquire and effectively use a large body of knowledge that is shared with others
- reason in ways which might be accepted by others using a shared, but implicit, common knowledge
- talk to others and make them understand our intention and meaning

All of these characteristics (and many more) are explained by the "Social Intelligence Hypothesis" (SIH) [18][6]. The SIH seeks to explain the evolutionary advantage provided by human brains. The explanation can be summarised as follows: our brains give us the social ability to coordinate and develop social knowledge basis and behaviours; this allows groups of individuals to inhabit specialized ecological niches (e.g. to live in the Tundra or Kalahari [20]); this ability to collectively adapt to and successfully inhabit a variety of niches gives us selective advantage. Under this view our brain is an adaptive organ to give us social abilities, what might be called *social intelligence*, including the characteristics listed above. However it also implies that these abilities are primary and classic displays of intelligence (including reasoning, and problem solving) are by-products of our social intelligence. Being able to solve a Rubrick's Cube or play chess has no selective advantage[7], but being part of a society that invents such puzzles and games and talks about them is!

---

[4] There are non-procedural ways of defining computable functions, e.g. using lambda calculus.
[5] Each intermediate state is the result of a different computation, but this is not the same as the one an intermediate state is part of unless it happens to be the final state.
[6] The related idea of "Machiavellian intelligence" had been around in primatology (e.g. [7]) before it was taken up in anthropology, focussing on the cognitive 'arms race' that might have occurred in terms of the competitive evolution in the social ability to make alliances [2]. The SIH is more general in its formulation covering a broad range of social abilities.

6 **Bruce Edmonds and Carlos Gershenson**

## 4. Design and the Turing Test

It is interesting that Turing chose a *social* test for intelligence years before the social roots of intelligence were widely appreciated. However, these social roots of human intelligence have implications for passing the TT. In particular it indicates that being part of the appropriate human culture is a key part of social intelligence, and not just an 'extra' that needs to be added once individual intelligence is sorted out. If this is the case, a considerable period of acculturation within a human society is necessary to pass the TT[8]. A purely individual or putative generalised intelligence without such a period of training would not suffice.

If a significant amount of learning is necessary to obtain to a social intelligence that might pass the TT, then to a large extent, this intelligence is not *designed* but developed in context. In other words, what one intentionally puts into a TM, i.e. by design, is not sufficient to pass the TT, but rather the huge body of context-specific information that is usually accumulated by humans as they grow into a culture. This is what we mean when we say that an "out-of-the-box" TM will not pass the TT.

Of course, once cultural information has been learnt by some entity, this could be 'compiled' into a complex TM, but we would not have *designed* this TM in any meaningful sense. However, this compiled entity is now a static entity and has stopped learning. Such a compiled TM might pass muster in a one-shot and short TT. However, if the test was extended in time then the conversation itself would form part of the cultural information that needs to be learnt about by the entity, so that in subsequent interactions it can appropriately refer back to what has been said. The difference is made clear if one has a conversation with a human with an impaired ability to lay down new memories, but who retains all the long-term memories before a certain date[9]. For the first period of time such people seem normal, but over any period of time their lack of memory becomes apparent[10].

---

[7] It is not even effective as a display for attracting a potential mate, as those on the back row of my undergraduate mathematics lectures demonstrated.

[8] Indeed, anecdotal accounts have it that Turing joked that such machines might be teased whilst they attended human schools.

[9] That such conditions exist and can be brought on by many different causes can be seen in many clinical accounts, e.g. [10].

[10] Of course, it might be that such a person might pass the TT if people with such conditions were expected as a possible participant. However, they would be clearly distinguishable from a healthy adult human.

## 5. Intelligence in General

It should be obvious that the sort of intelligence that could pass the TT requires reasoning *and* learning, it cannot be reasoning alone[11]. One consequence of this, along with the "No Free Lunch" theorems mentioned above [25], is that there is no such thing as *general* intelligence. That is to say that (unlike computation) intelligence cannot be general. Rather that different intelligences are each suited to particular problems and/or environments. Clearly, if the SIH is true, our intelligence is particularly suited to living in groups that develop a collective body of knowledge, habits, norms, skills, stories etc. Whatever "clever" algorithms we invent, using meta-reasoning, meta-learning, special cases, etc. will have "blind spots" where its algorithms are not as effective (w.r.t. its goals) as another (and possibly simpler) algorithm. Any algorithmic elaboration will have downsides as well as new abilities. In other words, intelligence is relative to the goals and environment of an entity, it is not an ordinal quality, with humans having the most.

Once one accepts that intelligence cannot have an ideal, we can distinguish different types of intelligence. For example, rats can be smarter than humans at navigating mazes. Does this imply that rats are smarter than humans? Well, it is better to clarify what is being tested. In the TT, human social interaction is being tested, humans tend to do pretty well. But different types of tests allow us to explore different types of intelligence, in animals and machines. For example, social insects can be pretty smart at collective decision making [11]. Plants could also be said to be intelligent [21], as well as bacteria [2][12]. We could take a narrow definition of intelligence and apply it to humans, or embrace a broad diversity of intelligences and understand human intelligence better by relating it to other types. For example, Randall Beer defines intelligence as "the ability to display adaptive behaviour" [1].

The same applies for artificial intelligence: we can take the narrow path of attempting to program human intelligence, which we argue is not achievable without learning/adaption; or we can take the broad path of exploiting all types of biological and artificial intelligences for solving problems in the most diverse areas. This broad approach is essential for our complex world. Given the fact that problems are changing constantly in unpredictable ways due to their interactions [13], systems we build to solve these problems must adapt and learn constantly, matching the timescale at which problems change. We can predict that, as complexity of problems increases, artificial systems will focus more and more on learning and adaption rather than on directly programming static solutions.

None of this invalidates the TT that, if passed, would unequivocally establish that a substantial intelligence, as impressive as human intelligence, had been brought into being. We would be as forced to recognise its legitimacy since we impute intelligence in our fellow humans on a similar basis.

---

[11] Reasoning is roughly coincident to computation, since any formal reasoning can be implemented as a computation and any computation as formal reasoning.

[12] Though it might be more accurate to attribute the intelligence in many cases to the process of evolution (the underlying *learning* process) rather than the entity that results.

## 6. Understanding Ourselves

Understanding human intelligence is obviously hard, due to its complexity. However, there is another difficulty as well – it's a "touchy" subject for us humans. We seek to understand ourselves using whatever cognitive means are at our disposal. What we use for this purpose has a strong influence upon our self-image.

The first and obvious way of understanding ourselves is by observing and understanding others around us. Whilst we clearly understand others by imagining how we would feel or think in their situation [6], it seems to also be true that, during development, we use our observations of how others behave to help us understand ourselves [15]. The circular bootstrapping of our own identity produces a very strong association between our perception of our own intelligence and that of others. It is perhaps this that makes the TT so compelling: we cannot but consider the entities with whom we converse as similar to ourselves.

A second way of understanding ourselves is by using analogies with devices around us. Whilst the Victorians might have conceived of the world and persons in terms of intricate clockwork, we tend to use the computer. Thinking of ourselves as computers is natural, since we have many things in common with modern computational devices: they are interactive, responsive and can be programmed to behave in human-like ways, for example learning. We can understand what computers are doing because, with practice, we can perform simple calculations and work out what small bit of computer code are doing. Intensive interaction with computers can lead to a strong association with such devices, to the extent that we see ourselves as a kind of computer and impute many human characteristics upon them [24]. Since the TM is a formal model of a computation, and being able to be predicted by such a model is what makes a device a "computer", this leads to an association of TMs and our intelligence. However, as the above arguments show, this analogy captures only some of the complete picture.

A third way of understanding ourselves, is as the pinnacle of evolution, as possessing (essentially) a completely general intelligence. Somehow it is assumed that the human brain, together with its creations: paper, maths, computers etc., are not limited in what it can understand. This is not a view of any individual, of course, but rather that eventually humankind will be able to work anything out. Surely, if people indeed think this, this is nothing more than sheer hubris. However forms of this thinking seem to be implicit in some of the thinking about the TT, e.g. French's [10] denigration of 'subcognitive' aspects of human intelligence and his criticism of the TT as "only" being a test of human intelligence.

We (along with many others) wish to work towards a different understanding of intelligence, by looking to other parts of life, their ecology and the process of evolution. This tries to relate the nature of intelligence to why it evolved. It is a more interactive view that wishes to place individual entities within a broader web of interactions, so that the interactions within an entity are just part of this broader web. It is admitted that, currently, this view does not have formal models of the strength of the TM, but rather a plethora of simulations and approaches.

## 7. Conclusions

There are good and general models of computation, going back to Turing's original paper but there is no equivalent for the interactive and continual process that we call learning, let alone the more general phenomena of adaptive interaction. The TT nicely shows the difference between the two, being a test that requires the latter. Perhaps Turing's later paper will have the effect of moving on thinking about ways of formally representing intelligence, and free it from the limited analogy of the TM.

## 8. Acknowledgements

BE was partially supported by the Engineering and Physical Sciences Research Council, grant number EP/H02171X/1. CG was partially supported by SNI membership 47907 of CONACyT, Mexico.